%% file: main.tex
\documentclass[letterpaper, 10 pt, conference]{ieeeconf}
\IEEEoverridecommandlockouts
\usepackage{graphicx, multirow}
\usepackage{mathptmx}
\usepackage{amsmath}
\usepackage{dsfont}
\usepackage{amssymb}
\usepackage{adjustbox}
\usepackage{wasysym}
\usepackage{subcaption}
\usepackage{nicefrac}
\usepackage{xspace}
\usepackage{comment}
\usepackage{booktabs}
\usepackage{arydshln}
\usepackage[table,x11names]{xcolor}
\usepackage[hidelinks]{hyperref}

\makeatletter
\DeclareRobustCommand\onedot{\futurelet\@let@token\@onedot}
\def\@onedot{\ifx\@let@token.\else.\null\fi\xspace}

\makeatother

\newcommand{\myparagraph}[1]{\vspace{4pt}\noindent\textbf{#1}}
\newcommand{\R}{\mathbb{R}}
\newcommand{\E}{\mathbb{E}}
\newcommand{\D}{\mathcal{D}}
\newcommand{\loss}{\mathcal{L}}

\usepackage{my_ref}

\title{\LARGE \bf
FedDrive: Generalizing Federated Learning to Semantic Segmentation in Autonomous Driving
}

\author{Lidia Fantauzzo$^{*,1}$, Eros Fan\`i$^{*,1}$, Debora Caldarola$^1$, Antonio Tavera$^1$, \\ Fabio Cermelli$^{1,2}$, Marco Ciccone$^1$, Barbara Caputo$^1$%
\thanks{$*$ Equal contribution.}
\thanks{$^1$ All the authors are supported by Politecnico di Torino, Turin, Italy. Corresponding authors: {\tt\scriptsize lidia.fantauzzo@studenti.polito.it}, {\tt\scriptsize eros.fani@polito.it}. Other authors: {\tt\scriptsize name.surname@polito.it}.}%
\thanks{$^2$ Fabio Cermelli is with Italian Institute of Technology, Genoa, Italy.}
}

\begin{document}

\addtolength{\topmargin}{.05in}
\pagenumbering{gobble}

\maketitle
\thispagestyle{plain}
\pagestyle{plain}

\begin{abstract}
Semantic Segmentation is essential to make self-driving vehicles autonomous, enabling them to understand their surroundings by assigning individual pixels to known categories.
However, it operates on sensible data collected from the users' cars; thus, protecting the clients' privacy becomes a primary concern.
For similar reasons, Federated Learning has been recently introduced as a new machine learning paradigm aiming to learn a global model while preserving privacy and leveraging data on millions of remote devices. 
Despite several efforts on this topic, no work has explicitly addressed the challenges of federated learning in semantic segmentation for driving so far. To fill this gap, we propose FedDrive, a new benchmark consisting of three settings and two datasets, incorporating the real-world challenges of statistical heterogeneity and domain generalization.
We benchmark state-of-the-art algorithms from the federated learning literature through an in-depth analysis, combining them with style transfer methods to improve their generalization ability.
We demonstrate that correctly handling normalization statistics is crucial to deal with the aforementioned challenges. Furthermore, style transfer improves performance when dealing with significant appearance shifts.
Official website: \href{https://feddrive.github.io}{https://feddrive.github.io}.
\end{abstract}

\input{sections/1-Introduction}

\input{sections/2-RelatedWork}
\input{sections/3-Benchmark}
\input{sections/4-Method}

\input{sections/5-Experiments}
\input{sections/6-Conclusions}
\input{sections/7-Acknowledgements}

\addtolength{\textheight}{-2.2pt}
\bibliographystyle{IEEEtran.bst} 
\bibliography{IEEEabrv,bibfile}
\end{document}

%% file: sections/1-Introduction.tex
\section{Introduction}
Research in autonomous driving aims at improving our safety and driving experience. 
In order to derive and execute trustworthy actions autonomously, vehicles must be able to perceive and understand their surroundings \cite{auto1, auto2, auto3, auto4, Fontanel_2021_CVPR}. To ensure the safety of the passengers, an autonomous vehicle needs to determine precisely whether there is a danger such as an obstruction or a pedestrian, and consequently decide whether to slow down or accelerate.
To do so, vehicles rely on the semantic segmentation task \cite{chen2017deeplab, xie2021segformer, oliveira2016efficient, bisenetv2}, whose goal is to provide a semantic prediction for every pixel of the image, giving a deep
understanding of the surrounding environment.
However, training robust semantic segmentation models requires having access to large scale datasets possibly representing all the conditions that can be encountered in the real world.
One possible solution is to collect images from the customers' vehicles which, being already on the road, face different situations and cover multiple geographic locations, weather conditions, viewpoints, etc. 
While collecting the images from vehicles is an effortless process, sending them to a central server to train a model could potentially violate the users privacy. Indeed, other nodes of the communication infrastructure may access personal and privacy-protected information, thus violating the regulations in force \cite{ad1,ad2,ad3}.

\begin{figure}[t]
\centering
\includegraphics[width=1.0\columnwidth]{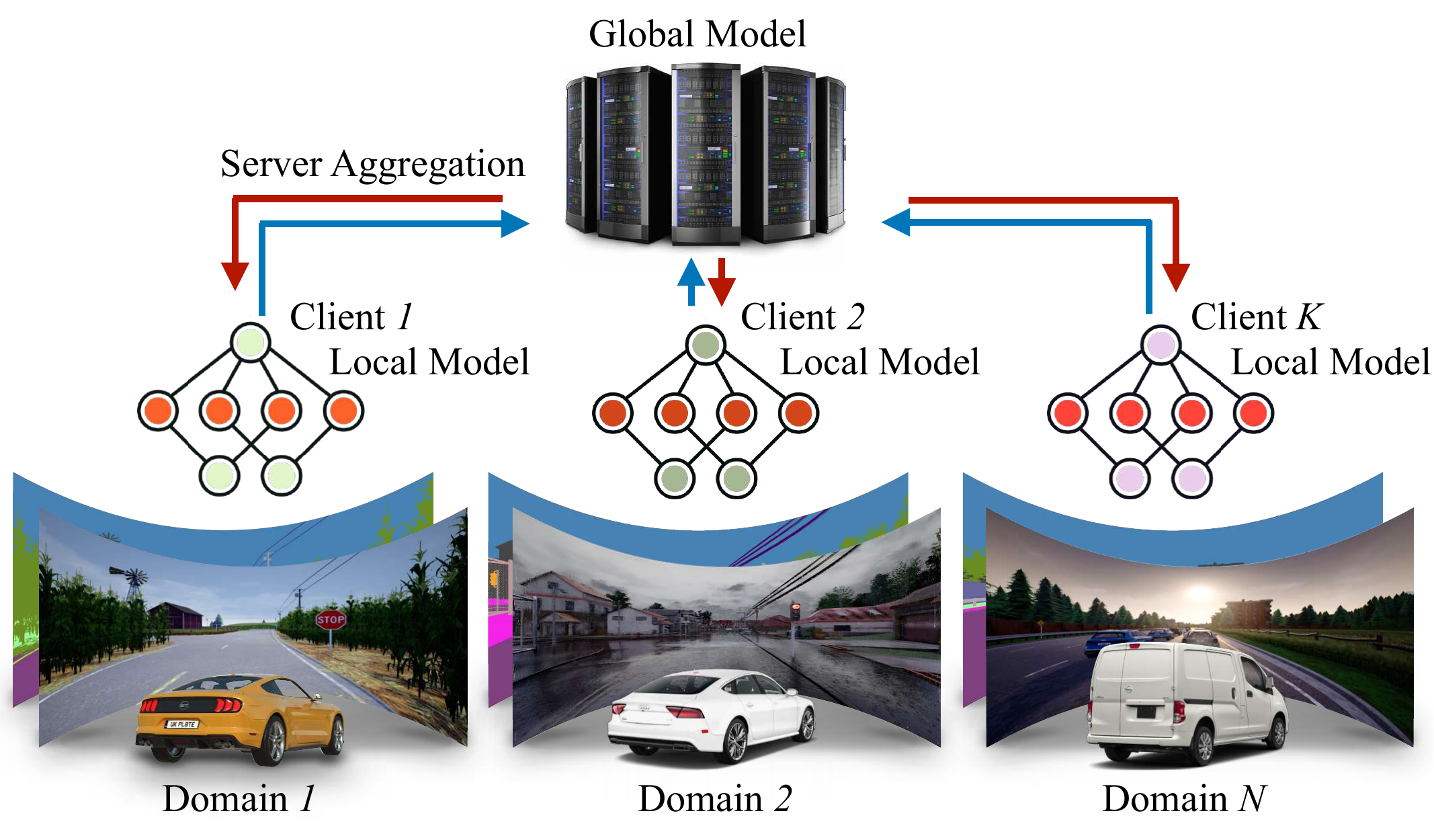}
\caption{
Multiple $K$ vehicles (\textit{clients}), driving in different $N$ domains, interact with a central \textit{server} to learn a global model while keeping route data private. At each communication round, a subset of clients performs training on its non-iid data, sharing only the local update with the server, which subsequently sends back the newly aggregated model.}
\label{fig:teaser}
\vspace{-5pt}
\end{figure}

A clever solution to train a model using all the clients' data while protecting their privacy is introduced by federated learning (FL) \cite{mcmahan2017communication}. FL algorithms learn a shared model leveraging a distributed training between several remote clients using their data in a privacy-compliant way.
However, FL is a relatively new field and current methods mainly focus on simple vision tasks, such as image classification \cite{mcmahan2017communication, hsu2020federated, SiloBN, FedBN}. Hence, the usage of FL for Semantic Segmentation in realistic urban environments opens the door to issues relating to the variety of witnessed scenes: images captured outdoors suffer from enormous variability in light, reflections, points of view, atmospheric conditions and types of settings, resulting in various \textit{domains} (\cref{fig:teaser}).
It is unclear how the current FL methods may perform under such a challenging scenario.

To answer this open question, we introduce the first Semantic Segmentation benchmark in a Federated Learning scenario applied to autonomous driving, called \textit{FedDrive}. 
While designing FedDrive, we focused on two important challenges that may occur when training a federated model in the real world: i) high statistical heterogeneity \cite{li2018federated, hsu2019measuring, heterog}, and ii) domain generalization \cite{SiloBN, FedBN}.
Statistical heterogeneity is a relevant issue for FL methods in real-world scenarios~\cite{hsu2020federated,heterog}. Each client (vehicle) usually observes part of a meta-distribution or different factors of variation (geographical location, viewpoint, weather, illumination), which are generally different from other clients. When increasing the heterogeneity of the clients, FL algorithms suffer in terms of convergence speed, increasing training time and hampering the final model performance.
On the other hand, Domain Generalization (DG) requires a model to perform well also on unseen conditions, \textit{i.e.} on domains different from the training ones. This is an essential ability for a model to operate in the real world, since not all the conditions can be anticipated at training time and a model should guarantee to safely operate on them.

To assess if current methods are able to deal with these real world challenges, we benchmark on FedDrive the main state-of-the-art federated learning algorithms FedAvg~\cite{mcmahan2017communication}, SiloBN~\cite{SiloBN}, FedBN~\cite{FedBN}, and different server optimizers~\cite{hsu2019measuring, server_optim}. Moreover, in order to deal with the domain shift, we combine them with style-transfer methods \cite{FFDG, lab} to obtain a model capable of effectively generalizing on unseen domains.

To summarize, our main contributions are the following:
\begin{enumerate}
    \item We introduce FedDrive, the first benchmark for FL in Semantic Segmentation applied to autonomous driving, focusing on the real world challenges of statistical heterogeneity and domain generalization. 
    \item We compare state-of-the-art federate learning methods and combine them with style transfer techniques to improve their generalization.
    \item We show the importance of using the correct clients' statistics when dealing with different domains and that style transfer techniques improve the performance on unseen domains,
    proving to be a solid baseline for future research in federated semantic segmentation.
\end{enumerate}

%% file: sections/2-RelatedWork.tex
\section{Related Work}
\subsection{Semantic Segmentation}
Semantic segmentation is a crucial task for autonomous driving applications whose goal is to predict the class of every pixel in the image. State-of-the-art methods employ an encoder-decoder architecture based on Convolutional Neural Networks \cite{chen2017deeplab, oliveira2016efficient} or Transformer modules \cite{xie2021segformer, cheng2021maskformer}. 
The vast majority of the segmentation methods assume to operate in a centralized setting, where the training is performed on a server and then deployed to multiple clients. To the best of our knowledge, no prior works investigated the real-world scenario of federated learning in semantic segmentation for autonomous driving.
It has been mostly addressed in the field of medical images \cite{ShellerMicahJ2019MDLM, LiWenqi2019PFBT, YiLiping2020SAEE, BerceaCosminI2021FDFL}, where these works only tried to solve the problem related to safeguarding the privacy of patients while using images collected from multiple sources, providing ad-hoc techniques.
Only recently, FedProto \cite{fedproto} proposed a general method to federated learning for object segmentation. They introduced a novel approach that computes client deviation using margins of prototypical representations learned on distributed data and applies them to drive federated optimization via an attention mechanism.

\subsection{Addressing Statistical Heterogeneity in FL}
Recent years have seen a growing interest in Federated Learning and its applications to the real world, focusing many efforts on finding an effective solution in the face of the marked heterogeneity of clients' data \cite{li2019convergence,heterog,li2020federated}.

The primary federated optimization approach FedAvg \cite{mcmahan2017communication} exploits a weighted average of the clients' updates for learning the global model. Unfortunately, while effective on homogeneous scenarios, FedAvg usually loses in convergence performance and speed when facing non-i.i.d. data \cite{li2019convergence,hsu2019measuring}. Many works try to deal with the \textit{client drift} introduced by the different data distributions by acting on the local optimization, so that the local models do not move in opposite directions with respect to the global one \cite{fedprox,karimireddy2020scaffold,hsu2020federated,acar2021federated}. Another line of research looks at the model aggregation introducing server-side momentum \cite{hsu2019measuring}, and optimizers \cite{server_optim} to deal with FedAvg lack of adaptivity. While proving effective even in the most challenging scenarios, all those works mainly focus on the heterogeneity based on \textit{labels} distribution skew. So far, there are only a few attempts at taking into account the non-i.i.d.ness induced by \textit{domain features} shift, which is of interest for our work. Following this direction, FedRobust \cite{FedRobust} addresses the affine distribution shift by learning affine transformations. Another consistent line of research tackles such issue by learning domain-specific batch normalization (BN) statistics \cite{SiloBN,caldarola2021cluster,FedBN}.
In particular, SiloBN \cite{SiloBN} keeps all BN statistics strictly local and leverages Adaptive Batch Normalization \cite{AdaBN} to address new domains at test time, while FedBN \cite{FedBN} introduces local BN layers. 
Both these methods result in personalized local models. Differently, we aim at learning a \textit{global} model capable of generalizing to different domains.

\subsection{Federated Domain Generalization}
Domain Generalization (DG) \cite{blanchard2011generalizing} seeks to extract a domain-agnostic model that can be applied to a previously unknown domain. Some strategies attempt to reduce domain shift between different source domains \cite{dg1, dg2}, while others rely on meta-learning solutions \cite{dg3, dg5, dg4}. Nonetheless, both solutions violate user privacy due to the necessity to access all data.

Methods that respect user privacy, such as \cite{dg6, dg7}, and \cite{dg8}, rely on deep neural networks, training heuristics, or data augmentation, respectively. Despite this, they cannot deal with imbalances in data distribution across domains.
Quande \textit{et al.} \cite{FFDG} were the first to present a solution to the DG problem in Federated Learning. They propose a Continuous Frequency Space Interpolation (CFSI) to exchange data distribution information among clients while protecting their privacy. In addition, \cite{medicalFDG} proposes a novel gradient alignment loss based on the Maximum Mean Discrepancy (MMD).
Following the footsteps of \cite{FFDG} and \cite{medicalFDG}, we show and analyze how state-of-the-art image-to-image translation methods, such as the frequency exchange in \cite{FFDG} or the LAB color space transformation proposed in \cite{lab}, can be used to address the problem of generalization to unknown domains in Federated Learning.

%% file: sections/3-Benchmark.tex
\input{tables/setting}

\section{FedDrive: A New Benchmark}

In this Section, we first formalize the federated scenario and the addressed task, explaining a fundamental issue arising in real world environments, \textit{i.e.} the data \textit{statistical heterogeneity} and the \textit{domain generalization}, and how it affects the autonomous driving setting. Then, we introduce FedDrive, the first benchmark for studying autonomous driving in federated scenarios.

\subsection{Federated Learning Setup}
The standard federated scenario is based on a client-server architecture. The server is the trusted party, while clients can be any computation-capable device, from personal smartphones to self-driving cars. Given a set of clients $\mathcal{K}$, with $|\mathcal{K}|=K$, a client $k \in \mathcal{K}$ has access to a \textit{privacy-protected} local dataset $\D_k$ of dimensions $n_k$. In the proposed scenario, each instance of $\D_k$ is an image $x\in \mathcal{X}$ made of $N$ pixels associated with its ground truth vector $y\in \mathcal{Y}^N$, denoting the set of $N$ tuples with elements belonging to the class space $\mathcal{Y}$.
Each pair is drawn from a local distribution $(x,y)\sim  P_k$. In autonomous driving scenarios, data distributions could be determined by the city in which the user lives, his usual driving routes, or typical weather conditions. 
The goal is to learn a global shared model $F(w):\mathcal{X} \rightarrow \R^{N\times |\mathcal{Y}|}$ with parameters $w$, mapping the image space $\mathcal{X}$ to a pixel-wise class probability vector, without breaking the privacy constraints. We aim to solve the task of assigning a semantic class $y_i\in\mathcal{Y}$ to each pixel $x_i$ of the input image $x\in\mathcal{X}$. That is accomplished through a learning paradigm based on $T$ communication rounds. During each round $t$, \textit{i)} the server exchanges the current model parameters $w^t$ with a subset $\mathcal{K}' \subseteq \mathcal{K}$ of the online devices, \textit{ii)} the selected clients train $F(w^{t})$ on their local data for $E$ epochs and send back the update $w^{t+1}_k$, \textit{iii)} all the updates are then aggregated on the server-side according to a chosen algorithm, consequently updating the global model. 

\subsection{Statistical Heterogeneity: A Real World Issue}
When dealing with realistic scenarios, users' data distributions may drastically differ. Looking at the world of autonomous driving, we can easily imagine how the data acquired from a car in Northern Europe will contain snow with a higher probability than those obtained from another one located on the Southern Coast. Alternatively, they may have access to similar weather conditions but different landscapes, \textit{e.g.} cityscapes instead of rural areas. These differences are brought together under the term \textit{statistical heterogeneity} and represent one of the main challenges faced by the FL community today \cite{hsu2019measuring,heterog,li2020federated}. Indeed, the standard federated paradigm is usually effective in homogeneous scenarios but fails at generalizing when facing heterogeneous - and more realistic - ones. Formally, given two clients {$(i,j)\in \mathcal{K}\times\mathcal{K}$} it is highly likely that $P_i \neq P_j \: \forall i \neq j$, where $P_i$ is the probability distribution associated to the $i$-th client. As a consequence $F_i(w) \neq F_j(w) \neq F(w)$, \textit{i.e.} the more the distributions are different, the greater the difference between locally learned models and the global one \cite{karimireddy2020scaffold}.
In general, let $(x,y)\sim P_i(x,y)$ be a sample drawn from the $i$-th client's local distribution. We usually face two kinds of heterogeneity:
i) \textit{label} distribution skew, \textit{i.e.} the marginal distribution $P_i(y)$ may vary across clients $i$ and $j$ even if $P_i(y|x) = P_j(y|x)$; ii) \textit{feature} distribution skew, \textit{i.e.} the same label $y$ can be associated with different features $x$ across clients. Formally, $P_i(x)$ and the conditional distribution $P_i(x|y)$ vary even if $P_i(y|x) = P_j(y|x)$ and $P_i(y)$ is shared, $\: \forall i\neq j$. For example, the same concept of house can be associated with totally different structures in the world. In this work, we focus on the latter and refer to the distinct feature distributions as \textit{\textbf{domains}}. 
The introduction of SS in urban environments exacerbates these issues due to the presence of a larger number of classes within a single image, which results in more possible distributions. Moreover, the outdoor scenes typical of the autonomous driving setting introduce further shifts due to changes in light, reflections, point of view, atmospheric conditions, etc. The presence of different domains in the peripheral devices will be referred to as \textit{domain imbalance}.

\subsection{Federated Domain Generalization}
In a realistic setting, not only clients have their own local data distribution, but new clients may appear at any time, introducing novel distributions. For example, in an autonomous driving scenario new clients may appear in a different location (such as in a different city or country) or may introduce a new vehicle with other characteristics. A robust model should be able to generalize to novel, unseen conditions to operate in the real world.
For this reason, in our benchmark, we focus on model generalization and explicitly assess a model's ability to operate on unseen domains at test time. 
Formally, evaluation will be performed on a test client $s$ having a distribution $P_{s}(x) \neq P_k(x) \: \forall k \in \mathcal{K}$. 

\subsection{FedDrive}
To investigate the different challenges of Federated Learning in autonomous driving, we introduce a new benchmark based on Cityscapes \cite{Cityscapes}, and IDDA \cite{idda} datasets. We propose multiple federated versions for each of them, based on different levels of data heterogeneity between clients. Table \ref{tab:fed_set} summarizes the multiple experimental settings emulating realistic scenarios.

\myparagraph{Cityscapes} is one of the most popular datasets for semantic segmentation and is a set of real photos taken in the streets of $50$ different cities with good weather conditions. The dataset contains $2975$ images for training and $500$ for testing, providing annotations for $19$ semantic classes.

The first and more na\"ive choice for adapting the dataset to the federated scenario consists of \textit{uniformly} splitting it among a fixed number of clients, \textit{i.e.} each image is randomly drawn and assigned to one of the users.

While this distribution allows for performance evaluation in a federated environment, it does not consider the real-world challenge of statistical heterogeneity. Therefore, we propose an additional split, denoted as \textit{heterogeneous}, that takes into account the information about the city where the photo was taken, resulting in a non-i.i.d. domain distribution across clients. Specifically, we set up 8 clients for each of the 18 training cities for a total of 144 devices and divide the images of each city among its clients.
The test client contains all the images belonging to cities never seen at training time for both the uniform and the heterogeneous settings.

\myparagraph{IDDA} is a synthetic dataset for semantic segmentation in the field of self-driving cars, providing annotations for 16 semantic classes with a broad variety of driving conditions, characterized by three axes: 7 towns (ranging from Urban to Rural environments), 5 viewpoints (simulating different vehicles), and 3 weather conditions (Noon, Sunset and Rainy scenarios) for a total of 105 domains.
As done for Cityscapes, we provide both a uniform and a heterogeneous version of the federated dataset. In the \textit{uniform} distribution, each client has access to 48 images
drawn randomly from the whole dataset. Since such distribution is highly unrealistic, we only use it as a reference for the model's performance. The \textit{heterogeneous} federated version of IDDA requires each client to see images belonging to a single domain. In addition, to test the generalization capabilities of the learned model, we use two test clients, one with images belonging to the already seen training domains (\textit{seen-dom}) and one with never seen ones (\textit{unseen-dom}). 

IDDA is further distinguished into two possible settings in order to analyze both \textit{semantic} and \textit{appearance} shift. {As for the former, the \textit{unseen-dom} test client contains images of a country town, while in the second case, the photos are taken in rainy conditions. We denoted this two setting variations respectively as \textit{country} and  \textit{rainy}.}

%% file: tables/setting.tex
\begin{table*}[t]
\caption{Summary of the settings in FedDrive. The uniform distribution is \textit{iid} while the heterogeneous is \textit{non-iid}.
}
\label{tab:fed_set}
\centering
\begin{tabular}{lccccc}
\toprule
\textbf{Dataset} & \textbf{Setting} & \textbf{Distribution} & \textbf{\# Clients}  & \textbf{\# Samples per user}& \textbf{Test clients} \\ \midrule
\multirow{2}{*}{Cityscapes} & - &  uniform & 146 & 10 - 40 &  unseen domains (new cities)\\ 
 & - &  heterogeneous &  144 & 10 - 45 &  unseen domains (new cities)\\ \midrule
\multirow{4}{*}{IDDA} & \multirow{2}{*}{country} & uniform & 90 & 48 & seen + unseen (country) domains  \\ 
 &  & heterogeneous & 90 & 48 & seen + unseen (country) domains  \\ \cmidrule{2-6}
 & \multirow{2}{*}{rainy} &  uniform & 69 & 48 & seen + unseen (rainy) domains  \\ 
 &  & heterogeneous & 69 & 48 & seen + unseen (rainy) domains \\
\bottomrule
\end{tabular}
\end{table*}

%% file: sections/4-Method.tex
\begin{figure*}[t]
\centering
\includegraphics[width=1.0\textwidth]{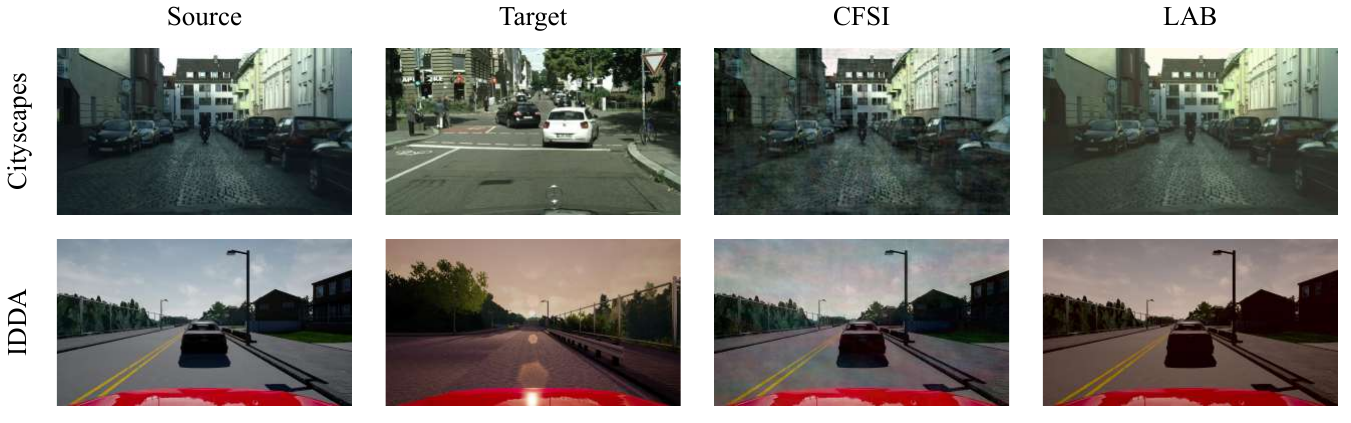}
\caption{
Examples of DG techniques used to transfer the domain from a source to a target image. CFSI produces artifacts if the amplitude interpolation tends toward the target; otherwise, the difference from the original image is not as noticeable. The LAB technique appears to work well in all conditions since the style of target is clearly transferred to the source.}
\label{fig:dg}
\end{figure*}

\section{Methods}

In this Section, we present the algorithms tested on our benchmark. Starting from the standard FedAvg \cite{mcmahan2017communication} and its adaptive version \cite{server_optim}, we address the domain shift between clients applying methods proper to the literature of heterogeneous Federated Learning \cite{SiloBN,FedBN}, Domain Adaptation and Generalization \cite{blanchard2011generalizing}, which are here described.  

\subsection{FedAvg: An Overview}
FedAvg \cite{mcmahan2017communication} is the standard approach for handling federated baselines. At each round $t$, given a subset $\mathcal{K}'$ of clients with $|\mathcal{K}'|=K'$ and the current global model parameters $w^t$, the goal is to optimize
\begin{equation}
    \arg\min_w \sum_{k=1}^{K'} p_k L_k(w)
\end{equation}
where $L_k = \E_{(x,y) \sim \D_k} \big[ \loss\big(F(w; x), y\big) \big]$ is the local empirical risk, $\loss$ is typically the cross-entropy loss and $p_k > 0$ is usually set to $\nicefrac{n_k}{n}$, with $n=\sum_k n_k$ being the total number of training images. The global model $w^{t+1}$ is then updated as $\nicefrac{1}{n}\sum_{k\in\mathcal{K}'} n_k w_k^t$. The authors of \cite{server_optim} show that applying FedAvg is equivalent to use SGD optimizer on the server-side with learning rate $\eta=1$. In particular, if $w_k^t$ is the $k$-th client's local update, we set $\Delta_k^t := w_k^t - w^t$ and $\Delta^t := \nicefrac{1}{K'} \sum_{k\in\mathcal{K}'} \Delta_k^t$. Then the FedAvg update corresponds to applying SGD to the pseudo-gradient $-\Delta^t$. Such formulation allows for testing different server-side learning rates and optimizers with the goal of improving FedAvg convergence. In \cref{sec:abl_server}, we propose an analysis using SGD with momentum \cite{hsu2019measuring}, Adam \cite{kingma2014adam} and Adagrad \cite{lydia2019adagrad}. 

\subsection{Batch Normalization for Domain Adaptation}
While FedAvg is very effective in homogeneous scenarios, it loses in performance when facing non-i.i.d. data. 
Previous works investigated the critical role of Batch Normalization \cite{IoffeSergey} layers in addressing non-i.i.d. data \cite{carlucci2017just,maria2017autodial,ManciniMassimiliano,AdaBN,tasknorm}.
Batch Normalization (BN) was introduced by Ioffe \& Szegedy \cite{IoffeSergey} to speed up the training with faster learning rates and reduce the sensitivity to the network initialization and have proved essential ever since. A BN layer receives as an input the batch $x = (x_1,\dots,x_B)$ of size $B$ with $x_i\in\mathcal{X} \: \forall i\in[B]$ and normalizes it as 
\begin{equation}
    BN(x) = \gamma \dfrac{x - \mu}{\sqrt{\sigma^2 + \epsilon}} + \beta
\end{equation}

where  $\mu$ and $\sigma$ are the \textit{BN statistics} that correspond to the running mean and variance of each channel, respectively; the scale $\gamma$ and bias $\beta$ are learnable parameters, and $\epsilon$ is a small scalar that prevents division by 0. 

The main drawback of BN layers is the assumption that training and test data are drawn from the same distribution \cite{tasknorm}, which makes them unsuitable for our scenario. 
As a solution, Li \textit{et al.} \cite{FedBN} propose FedBN, which leverages BN layers to alleviate the feature shift occurring during the server-side parameter aggregation. The local BN layers are updated locally but never shared and aggregated on the server-side, resulting in personalized parameters. The main issue arising with such an approach is the addressing of unseen domains or clients at test time.
SiloBN \cite{SiloBN} differentiates the roles played by the BN statistics $\{\mu,\sigma\}$ and the learnable parameters $\{\gamma,\beta\}$: the former capture \textit{local} domain information, while the latter are transferable across domains. That is why the clients only share the parameters, keeping the statistics local to address the heterogeneity. New clients and domains at test time are handled using AdaBN \cite{AdaBN}, recomputing the BN statistics on the new testing domain while freezing all the other parameters of the server model. Both FedBN and SiloBN will be used as baselines in  \cref{sec:exp}.

\subsection{Generalization through Style Translation}
SiloBN and FedBN address statistical heterogeneity by manipulating the BN statistics. However, in order to improve the generalization capabilities, some recently proposed techniques go beyond the model and directly modify the images \cite{FFDG, lab}. An example of these methods is reported in \cref{fig:dg}.
The fundamental idea behind these methods is to share distribution information among clients to overcome the limits of decentralized datasets. 

\myparagraph{Continuous Frequency Space Interpolation} (CFSI) \cite{FFDG}.
Given an image $x_i^k$ from the $k$-th client, the frequency space signal can be derived from the Fast Fourier Transform (FFT). The low-level information (\textit{i.e.} color, brightness, etc.) is reflected in the amplitude spectrum of the FFT. To share the distribution information among local clients, a distribution bank $\mathcal{A}= \{\mathcal{A}^1,\dots,\mathcal{A}^K\}$ is firstly created, where each $\mathcal{A}^k = \{\mathcal{A}_i^k\}_{i=1}^{n_k}$ contains all amplitude spectrum of images from the $k$-th client. A continuous interpolation of the frequencies is used to transfer multi-source distribution to each local client.
\cite{FFDG} interpolates each client image to each domain distribution. Still, due to the significant computation required by the style transfer methods, this is not scalable or feasible in a realistic setting with several domains. As a result, we apply CFSI transformation to only half of the client data and sample only one target amplitude spectrum randomly from the shared distribution bank $\mathcal{A}$. The interpolation ratio $\lambda$ is used to sample images uniformly in a range $[0, 1]$.

\myparagraph{LAB-based Image Translation} (LAB) \cite{lab}.
To share the distribution information across local clients, a shared bank $\mathcal{L}={\mathcal{L}^1,\dots,\mathcal{L}^K}$ is created, where each $\mathcal{L}^k = \{\mu_i^k, \sigma_i^k\}^{n_k}_{i=0}$ is the set of means and standard deviations for each image of client $k$, transformed in the LAB color space. As previously done, we translate only half of the client data. More specifically, an RGB image $x_S^{RGB}$ of a client $k$ is firstly converted to the LAB color space $x_S^{LAB}$. Then, the mean $\mu_S$ and the standard deviation are computed for each channel of $x_S^{LAB}$. A random pair of $(\mu_T, \sigma_T)$ is sampled from the distribution bank $\mathcal{L}$. The $x_S^{LAB}$ image style is then translated as follow: 
\begin{equation*}
    \hat{x}_S^{LAB} = \dfrac{x_S^{LAB}-\mu_S}{\sigma_S} \ast \sigma_T + \mu_T.
\end{equation*}
Following the alignment of the distribution, the translated LAB image $\hat{x}_S^{LAB}$ is converted back to the RGB color space as $\hat{x}_S^{RGB}$, for the subsequent training phase.

%% file: sections/5-Experiments.tex
\section{Experiments}
\label{sec:exp}
\subsection{Implementation Details}
To fully exploit the limited capabilities of local devices, we choose a BiSeNet V2 \cite{bisenetv2} architecture, a lightweight network with two principal branches capturing spatial features and high-level semantic context. For simplicity, we use the common vanilla FedAvg aggregation with SGD as server optimizer for all the experiments, with learning rate $1$. We also provide an ablation on the server optimizer, showing that adding momentum helps reaching very high performances in the uniform and heterogeneous settings and seen domain, while giving only few points of improvement in the domain shift case. We minimize the cross-entropy loss on every client with SGD as local client optimizer with an initial learning rate of $0.05$ for Cityscapes and $0.1$ for IDDA, momentum $0.9$, weight decay $0.0005$ for Cityscapes and no weight decay for IDDA, and batch size $16$. We point out that when the clients do not have enough samples to use 16 as batch size, we virtually double their dataset by horizontally flipping the images.
Following \cite{bisenetv2}, we use a polynomial learning rate schedule on every client, \textit{i.e.} the initial learning rate is multiplied by $(1 - \frac{\text{iter}}{\text{max\_iter}})^{0.9}$. We apply OHEM (Online Hard-Negative Mining) \cite{shrivastava2016training}, selecting 25\% of the pixels having the highest loss for the optimization.
We apply a standard data augmentation pipeline: we randomly scale the images in the range $(0.5, 1.5)$ for Cityscapes and $(0.5, 2)$ for IDDA, then extract a crop with size $512\times1024$ for Cityscapes and $512\times928$ for IDDA
Following a convergence analysis, we decide to use $5$ clients per round and $2$ local epochs as it is the optimal combination in terms of speed of convergence and overall runtime. In the lack of a previous reference, we set the total number of rounds for all the experiments at $1600$, 
We decide to report the mean and the standard deviation of the mean Intersection over Union (mIoU) of the evaluations every $5$ rounds over the last $100$ rounds due to some instability of the test results.

\subsection{Results}

\input{tables/cityscapes}

\begin{figure*}[t]
\centering
\includegraphics[width=.9\textwidth]{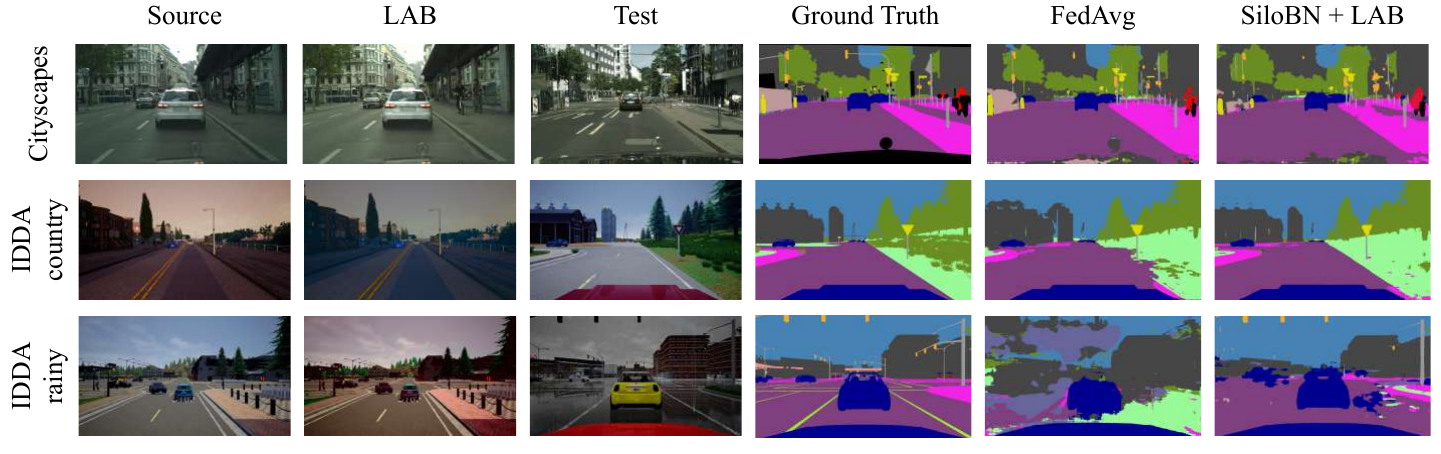}
\caption{
Qualitative results in the \textit{heterogeneous} scenarios for Cityscapes and for the two IDDA settings.}
\label{fig:qualitatives}
\end{figure*}

\myparagraph{Cityscapes.} \cref{tab:cts_fed} reports Cityscapes experiments.

The first line shows the performance of FedAvg in the \textit{uniform} scenario ($45.62 \pm 1.25\%$) as a reference for the performance in the absence of domain shift.

The experiments with style translation, especially the CFSI ones, record slightly higher standard deviations. In Cityscapes, the improvements brought by SiloBN are consistent (about 9 or 10 percentage points of mIoU in favor of the SiloBN experiments), despite the domain shift across cities being primarily related to semantics rather than style. Indeed, test images belong to cities never seen during training but to the same country and are taken in similar weather conditions. 

On the contrary, the style translation methods do not contribute to performance improvement: we recorded performance drops of about 3 and 1 percentage points of mIoU between the FedAvg experiment and the FedAvg + CFSI ($40.55 \pm 2.15\%$ vs $43.33 \pm 1.66\%$) and FedAvg + LAB experiments ($42.69 \pm 2.07\%$ vs $43.33 \pm 1.66\%$), respectively, while the performances are similar between no style translation, CFSI, and LAB, when applying SiloBN.

\myparagraph{IDDA.} 
In \cref{tab:idda_fed}, the experiments on the IDDA dataset are reported for the \textit{heterogeneous} distribution of both \textit{country} and \textit{rainy} settings.

Similarly to Cityscapes, the first row reports the performance of FedAvg in the uniform scenario as a reference. We can see that good performances are obtained when using the same domain test client (\textit{seen}). However, on the \textit{unseen} test client, the performances are lower, especially in the rainy setting, where the mIoU drops by 35 mIoU percentage points on average with respect to the \textit{seen} test client. This underlines the significant domain shift introduced by the \textit{unseen} test client.
We remark that country and rainy settings provide different challenges, as can be seen from \cref{fig:qualitatives}. The country domain has a similar style w.r.t. the training clients, since the images are taken in different weather conditions, but has other semantic characteristics, \textit{e.g.} it rarely contains sidewalks near the road, as instead frequently happens on the training clients. On the opposite, the rainy domain has similar semantic characteristics but an essentially different appearance since there are no rainy images on training.

When introducing a domain imbalance, the performance of FedAvg has a significant drop, especially in the seen test client. 
On average, the performance on the seen test client drops by nearly 23 mIoU percentage points, achieving 42.43\% on the country setting and 38.18\% on the rainy one. This result emphasizes the domain imbalance across training clients and the high statistical heterogeneity of this setting. Differently, the performance drop on the unseen test client is limited, with a drop of about 2 mIoU percentage points on the country and a drop of about 12 mIoU percentage points on the rainy setting.
\input{tables/idda}

The introduction of techniques able to cope with statistical heterogeneity largely improves the performance on the seen test client. FedBN and SiloBN show an improvement, on average, of 14 mIoU percentage points on country and of 21 mIoU percentage points on rainy settings, strongly reducing the domain gap between training clients. 
Moreover, analyzing the results on the unseen test client, we see two different behaviors of SiloBN. On the country setting, it shows only a limited improvement with respect to FedAvg: $40.01 \pm 1.26 \%$ FedAvg vs $45.32 \pm 0.90 \%$ SiloBN. Differently, it largely improves the performance on the rainy setting: $26.75 \pm 2.32\%$ FedAvg vs $50.03 \pm 0.79\%$ SiloBN. These results confirm that SiloBN strongly reduces the domain shift related to the appearance. Still, it does not alleviate much the semantic domain shift present in the country domain.

Overall, we can see that applying CFSI and LAB style transfer methods improve the performance across domains and settings. In particular, on the seen test client, LAB provides a considerable performance boost when applied to both FedAvg and SiloBN: on the country setting, it improves FedAvg by about 22 mIoU percentage points and SiloBN by about 6 percentage points, while on the rainy setting, it gains about 27 mIoU percentage points on FedAvg and about 4 mIoU percentage points on SiloBN. We note that it largely improves FedAvg, while it is less effective on SiloBN since it has an effect similar to SiloBN: it transfers the style of the training domains across clients' images. However, we note that the style transfer is applied to the images rather than the BN statistics, as in SiloBN.
When considering the performance on a different test domain (unseen), both LAB and CFSI improve the performance except in the rainy domain when applied to FedAvg, where LAB is less effective: FedAvg $26.75 \pm 2.32 \%$ vs CFSI $31.05 \pm 2.68 \%$ vs LAB $26.82 \pm 1.78 \%$.
Differently, we can see that LAB, when combined with SiloBN (\cref{fig:qualitatives}), improves performance in both the country and rainy settings, by around $5$ and $4$ mIoU percentage points respectively compared to SiloBN, and by about 10 and 27 mIoU percentage points compared to FedAvg, respectively.
Note how the combination of FedAvg + SiloBN + LAB is even much better than applying FedAvg only on the uniform split, expecially on the rainy unseen test domain, where it performs about 26 mIoU percentage points more than the uniform FedAvg only experiment.

Finally, observe how the standard deviation is generally higher for the unseen test domain than the seen test domain for both country and rainy tests, how it is higher for rainy than country experiments, underlining how the rainy partition is much more challenging than the country partition once again, and that CFSI and LAB contribute in reducing the variability of the results.

\subsection{Ablation study on server optimizer} \label{sec:abl_server}
In \cref{tab:idda_fed_server} we report the ablation study on the server optimizer on the IDDA settings.
We compared four server optimizers: SGD with learning rate 1 and no momentum, FedAvgM \cite{hsu2019measuring}, Adam \cite{kingma2014adam} and AdaGrad \cite{lydia2019adagrad}: for the last two we report the best result among the experiments with two learning rates: 0.1 and 1.0.
According to the table, FedAvgM clearly outperforms the other server optimizers: in the uniform experiments it increases the average performance by around 8 mIoU percentage points for both the country and rainy settings for the seen test domain, while it increases the average performance by around 6 and 2 mIoU percentage points for the country and rainy partitions for the unseen test domain, respectively. For the heterogeneous distribution the increase in performance is lower but still worth, ranging from 2 to 5 mIoU percentage points according to the particular experiment and setting. Finally, the benefits of FedAvgM are less visible when using SiloBN, with slightly improvements and even a small decrease in performance of about 1.5 mIoU percentage points for the unseen test domain in the rainy setting with respect to the SGD server optimizer. In all the experiments, Adam and AdaGrad are not able to significantly outperform SGD and to get closer to FedAvgM results.

\input{tables/optimizer}

%% file: tables/cityscapes.tex
\begin{table}[t]
    \caption{FedDrive: Cityscapes results.}
    \label{tab:cts_fed}
    \centering
    \begin{adjustbox}{width=0.75\columnwidth}
    \centering
    \begin{tabular}{lc}
    \toprule
        \multicolumn{1}{c}{\textbf{Method}} &  \textbf{mIoU $\pm$ std (\%)} \\
        \midrule
          \rowcolor{lightgray} FedAvg (uniform) & 45.62 $\pm$ 1.25\\
          FedAvg        & 43.33 $\pm$ 1.66 \\
          FedAvg + CFSI & 40.55 $\pm$ 2.15 \\ 
          FedAvg + LAB  & 42.69 $\pm$ 2.07 \\ 
          SiloBN        & 52.86 $\pm$ 1.29 \\
          SiloBN + CFSI & 52.11 $\pm$ 1.83 \\
          SiloBN + LAB  & \textbf{53.37 $\pm$ 1.65} \\
    \bottomrule
    \end{tabular}
    \end{adjustbox}
    \vspace{-5pt}
\end{table}

%% file: tables/idda.tex
\begin{table}[t]
    \caption{FedDrive: IDDA results in mIoU $\pm$ std (\%).}
    \label{tab:idda_fed}
    \centering
    \begin{adjustbox}{width=1.0\columnwidth}
    \centering
    \begin{tabular}{lcccc}
    \toprule
        &  \multicolumn{2}{c}{\textbf{country}} & \multicolumn{2}{c}{\textbf{rainy}}\\
        \textbf{Method} &  \textbf{seen} & \textbf{unseen} & \textbf{seen} & \textbf{unseen}\\\midrule
        \rowcolor{lightgray} FedAvg (uniform) & 63.57 $\pm$ 0.60 & 49.74 $\pm$ 0.79 & 62.72 $\pm$ 3.65 & 27.61 $\pm$ 2.80 \\
        FedAvg         & 42.43 $\pm$ 1.78 & 40.01 $\pm$ 1.26 & 38.18 $\pm$ 1.40 & 26.75 $\pm$ 2.32 \\ 
        FedAvg + CFSI  & 54.70 $\pm$ 1.12 & 45.70 $\pm$ 1.73 & 55.24 $\pm$ 1.65 & 31.05 $\pm$ 2.68 \\ 
        FedAvg + LAB   & 56.59 $\pm$ 0.90 & 45.68 $\pm$ 1.04 & 58.85 $\pm$ 0.89 & 26.82 $\pm$ 1.78 \\ 
        FedBN          & 54.39            & -                & 56.45            & - \\
        SiloBN         & 58.82 $\pm$ 2.93 & 45.32 $\pm$ 0.90 & 62.48 $\pm$ 1.42 & 50.03 $\pm$ 0.79 \\
        SiloBN + CFSI  & 61.22 $\pm$ 3.88 & 49.17 $\pm$ 1.01 & 63.04 $\pm$ 0.31 & 50.54 $\pm$ 0.88 \\
        \textbf{SiloBN + LAB} & \textbf{64.32} $\pm$ \textbf{0.76} & \textbf{50.43} $\pm$ \textbf{0.63} & \textbf{65.85} $\pm$ \textbf{0.91} & \textbf{53.99} $\pm$ \textbf{0.79} \\
    \bottomrule
    \end{tabular}
    \end{adjustbox}
    \vspace{-5pt}
\end{table}

%% file: tables/optimizer.tex
\begin{table}[t]
    \caption{Server optimizers comparison on IDDA.}
    \label{tab:idda_fed_server}
    \centering
    \begin{adjustbox}{width=1.0\columnwidth}
    \centering
    \begin{tabular}{lclcccc}
    \toprule
        & & & \multicolumn{2}{c}{\textbf{country (mIoU\%)}} & \multicolumn{2}{c}{\textbf{rainy (mIoU\%)}}\\ %
        & \textbf{Method} & \textbf{Optimizer} & \textbf{seen} & \textbf{unseen} & \textbf{seen} & \textbf{unseen}\\\midrule
        \multirow{4}{*}{\rotatebox[origin=c]{90}{uniform}} & 
        \multirow{4}{*}{FedAvg} & SGD & 63.57 $\pm$ 0.60 & 49.74 $\pm$ 0.79 & 62.72 $\pm$ 3.65 & 27.61 $\pm$ 2.80 \\ 
        &  & \textbf{FedAvgM} & \textbf{71.27} $\pm$ \textbf{0.85} & \textbf{55.47} $\pm$ \textbf{1.07} & \textbf{70.99} $\pm$ \textbf{0.71} & \textbf{29.83} $\pm$ \textbf{2.03} \\ 
        &  & Adam    & 63.31 $\pm$ 0.37 & 50.21 $\pm$ 0.40 & 65.39 $\pm$ 0.52 & 31.72 $\pm$ 1.74 \\
        &  & AdaGrad & 59.44 $\pm$ 0.94 & 46.09 $\pm$ 0.83 & 59.45 $\pm$ 1.30 & 27.80 $\pm$ 1.29 \\ \midrule
        \multirow{8}{*}{\rotatebox[origin=c]{90}{heterogeneous}} & 
        \multirow{4}{*}{FedAvg} & SGD & 42.43 $\pm$ 1.78 & 40.01 $\pm$ 1.26 & 38.18 $\pm$ 1.40 & 26.75 $\pm$ 2.32 \\ 
        &  & \textbf{FedAvgM} & \textbf{44.38} $\pm$ \textbf{1.98} & \textbf{42.42} $\pm$ \textbf{2.15} & \textbf{41.21} $\pm$ \textbf{1.98} & \textbf{31.91} $\pm$ \textbf{3.77} \\ 
        &  & Adam    & 39.93 $\pm$ 2.44 & 38.15 $\pm$ 1.89 & 37.97 $\pm$ 2.04 & 28.47 $\pm$ 2.56 \\
        &  & AdaGrad & 40.89 $\pm$ 2.05 & 38.03 $\pm$ 1.65 & 39.02 $\pm$ 1.71 & 27.08 $\pm$ 2.85 \\ \cmidrule{2-7}
        & \multirow{4}{*}{SiloBN} & \textbf{SGD} & 58.82 $\pm$ 2.93 & 45.32 $\pm$ 0.90 & 62.48 $\pm$ 1.42 & \textbf{50.03} $\pm$ \textbf{0.79} \\
        &  & \textbf{FedAvgM}  & \textbf{61.99} $\pm$ \textbf{1.51} & \textbf{46.20} $\pm$ \textbf{1.20} & \textbf{63.69} $\pm$ \textbf{1.25} & 48.49 $\pm$ 1.04 \\ 
        &  & Adam     & 58.36 $\pm$ 1.26 & 42.31 $\pm$ 0.84 & 61.51 $\pm$ 0.90 & 47.22 $\pm$ 0.89 \\ 
        &  & AdaGrad  & 48.97 $\pm$ 1.34 & 41.69 $\pm$ 1.31 & 54.06 $\pm$ 1.29 & 45.80 $\pm$ 0.89 \\
    \bottomrule
    \end{tabular}
    \end{adjustbox}
    \vspace{-5pt}
\end{table}

%% file: sections/6-Conclusions.tex
\section{Conclusions}
We presented FedDrive, a benchmark for Semantic Segmentation in Federated Learning suited for the autonomous driving task. We proposed a realistic set of real-world scenarios to investigate the impact of heterogeneous domain distributions across clients, taking into account the constraints posed by stylistic and semantic shifts.
We assess methods from both the Federated Learning and Domain Generalization literature to test our proposed benchmark and highlight the open challenges. We show how SiloBN methods contribute to performance improvement while DG methods such as LAB and CFSI do not contribute to or even worsen the final performance when dealing with the semantic shifts in Cityscapes.
In comparison, when the gap in IDDA concerned semantics and style, SiloBN had an even more positive impact, confirming its effectiveness only when the domain shift is related to appearance.
Finally, we show that SiloBN and LAB together increases performance in all IDDA scenarios, alleviating the issues in the face of diverse types of domain shifts.
We believe that FedDrive will foster research towards privacy-preserving autonomous driving.

%% file: sections/7-Acknowledgements.tex
\section{Acknowledgements}

This work was supported by the Consorzio Interuniversitario Nazionale per l'Informatica (CINI).